\documentclass[journal]{IEEEtran}
\usepackage{times}
\usepackage{epsfig}
\usepackage{graphicx}
\usepackage{amsmath}
\usepackage{amssymb}
\usepackage{multirow}
\usepackage{colortbl}
\usepackage{listings}
\usepackage{footnote}
\usepackage{bm}
\usepackage{cite}
\usepackage{booktabs}
\usepackage{url} 
\usepackage{algorithm}
\usepackage{algorithmic}
\usepackage{colortbl}
\usepackage{hyperref}
\usepackage[dvipsnames]{xcolor}
\usepackage{pifont}
\usepackage{tikz-cd}

\begin{document}
\title{FractalMamba++: Scaling Vision Mamba Across Resolutions via Hilbert Fractal Geometry}

\author{Bo Li~\IEEEmembership{Member,~IEEE}, Haoke Xiao, Lv Tang
\thanks{Bo Li is with vivo Mobile Communication Co., Ltd, Shanghai, China. Haoke Xiao is with Kuaishou Technology, Beijing, China. Lv Tang is with University of Alberta, Department of Electrical and Computer Engineering, Edmonton, Canada. Email: libra@vivo.com, luckybird1994@gmail.com.}}

\markboth{Submit to IEEE Transactions on Multimedia}%
{Li \MakeLowercase{\textit{et al.}}: FractalMamba++ for Resolution-Scalable Vision Mamba}

\IEEEtitleabstractindextext{
\begin{abstract}
Vision Mamba offers linear complexity for long visual sequences, yet its performance depends critically on how a two-dimensional patch grid is serialized into a one-dimensional state-space recurrence.
Raster-style scans disrupt spatial continuity, and the mismatch between 2D locality and 1D state propagation becomes increasingly severe when the inference resolution grows beyond the training grid.
This paper presents FractalMamba++, a resolution-scalable vision backbone organized around a single geometric principle: the recursive self-similar structure of the Hilbert curve determines how patches are serialized, where long-range state shortcuts are inserted, and how positional relations are encoded.
First, Hilbert-curve-based Fractal Serialization preserves local 2D neighborhoods more faithfully than linear scans and provides consistent neighborhood statistics across resolutions.
Second, the Fractal Hierarchy Skip Connection (FHSC) derives a compact set of deterministic state-injection routes from Hilbert recursion levels, mitigating long-sequence information fading without runtime search, hand-derived gradients, or dedicated CUDA kernels.
Third, Fractal-Aware 2D Rotary Position Encoding (FA-RoPE) combines normalized 2D coordinates with a fractal hierarchy level so that feature interactions depend on actual spatial proximity and recursive structural role rather than serialized 1D distance.
Extensive experiments on ImageNet-1K classification, COCO detection and instance segmentation, ADE20K semantic segmentation, and LEVIR-CD+ remote sensing change detection show that FractalMamba++ improves over existing Mamba-based vision backbones, especially under high-resolution inputs.
\end{abstract}
\begin{IEEEkeywords}
Vision Mamba, State Space Models, Fractal Serialization, Hilbert Curve, Fractal Hierarchy Skip Connection, Fractal-Aware Rotary Position Encoding, Resolution Scalability
\end{IEEEkeywords}}

\maketitle
\IEEEdisplaynontitleabstractindextext
\IEEEpeerreviewmaketitle

\section{Introduction}

\IEEEPARstart{F}{oundation} models~\cite{DBLP:conf/naacl/DevlinCLT19,DBLP:conf/icml/RadfordKHRGASAM21,DBLP:conf/icml/0001LXH22,DBLP:journals/jmlr/ChowdheryNDBMRBCSGSSTMRBTSPRDHPBAI23,DBLP:journals/corr/abs-2303-08774,DBLP:journals/corr/abs-2302-13971,DBLP:journals/corr/abs-2304-07193,Kirillov_2023_ICCV,DBLP:conf/icml/0008LSH23,DBLP:conf/eccv/ZhengJWZCWL24,ravi2024sam2,chen2024far,awais2025foundation,liu2025graph,DBLP:journals/pami/LiJHWZLMZ25} have become central to artificial intelligence, owing to their ability to learn generalizable representations through large-scale pre-training and to adapt to diverse downstream tasks~\cite{DBLP:conf/cvpr/LaiTCLY0J24,DBLP:journals/ijcv/TangJXL25,zhuang2025vargpt,DBLP:journals/corr/abs-2502-19634}. At the heart of most foundation models lies the Transformer architecture~\cite{DBLP:conf/nips/VaswaniSPUJGKP17}, whose self-attention mechanism is burdened by quadratic complexity in sequence length. This quadratic scaling becomes a severe bottleneck when the input is long, as in high-resolution images, and has motivated the search for alternative backbones. Mamba~\cite{DBLP:conf/icml/DaoG24,DBLP:journals/corr/abs-2312-00752} is one such alternative: by replacing attention with linear-time state-space recurrences, it achieves linear complexity for training and inference. A line of vision Mamba models~\cite{DBLP:journals/corr/abs-2403-09338,DBLP:journals/corr/abs-2403-17695,DBLP:conf/icml/ZhuL0W0W24,DBLP:conf/nips/LiuTZYX0YJ024,DBLP:journals/corr/abs-2406-02395} has adapted this idea to computer vision by scanning patches into a 1D sequence that a state space model (SSM) can process.

\begin{figure}[!t]
    \centering
    \includegraphics[width=\linewidth]{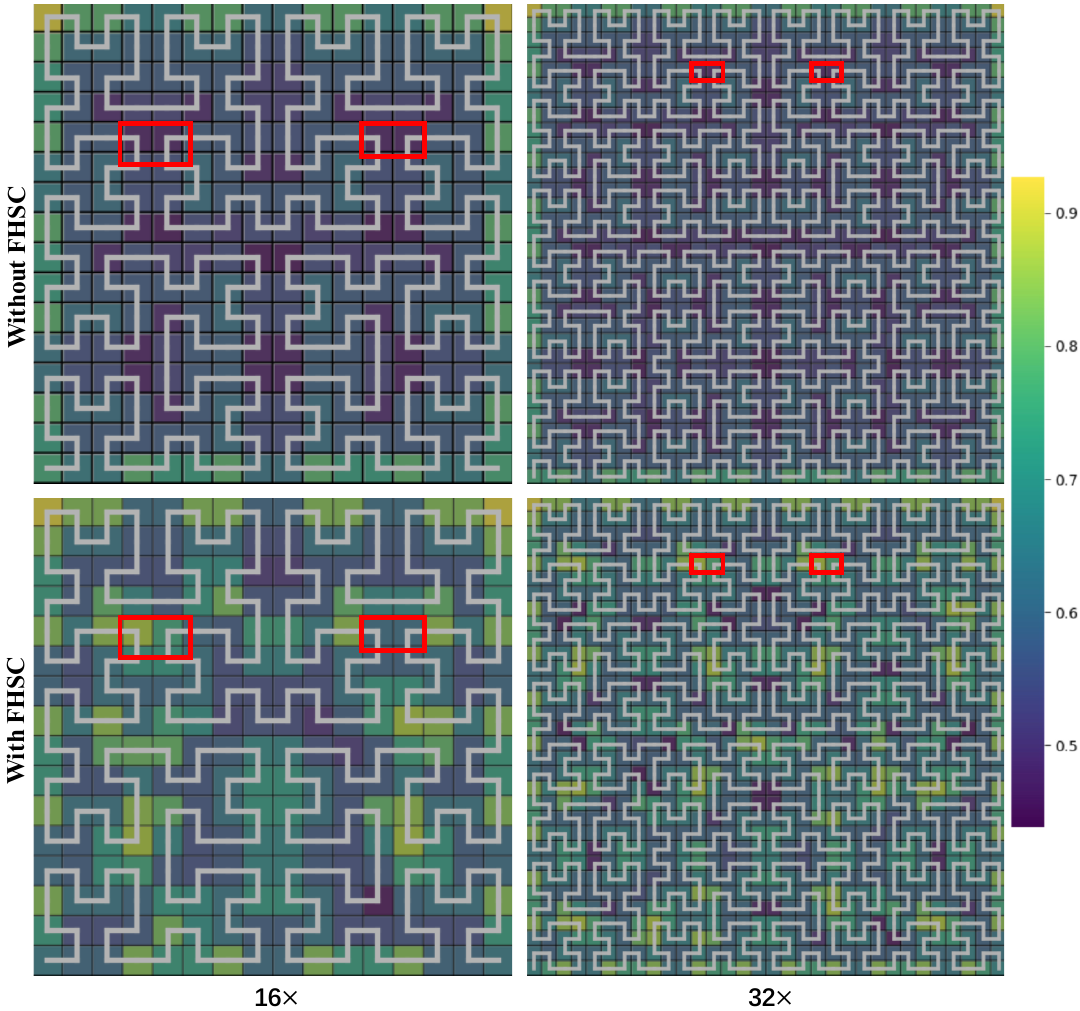}
    \caption{Visualization of correlation between the final state and state tokens in the deepest feature layer under different input resolutions. The $16\times$ and $32\times$ maps correspond to the output token grids for input resolutions of $512\times512$ and $1024\times1024$, respectively. Without FHSC, correlations are biased and localized due to information fading across long sequences. With FHSC enabled, stronger and more distributed correlations are observed, indicating improved global context propagation. Colors closer to \textcolor{yellow!70!black}{yellow} indicate higher correlations, while \textcolor{blue!60!black}{darker blue} colors indicate lower correlations.}
    \label{Introd_FHSC_Map}
    \vspace{-0.3cm}
\end{figure}

Despite this progress, the adaptation step remains fragile: structure is two-dimensional, whereas an SSM recurrence is one-dimensional. Converting a 2D patch grid into a sequence is not a benign detail but a modeling decision that determines which visual neighbors can exchange information through state transitions. Standard raster or row-major scans~\cite{DBLP:journals/corr/abs-2403-09338,DBLP:journals/corr/abs-2403-17695,DBLP:conf/icml/ZhuL0W0W24,DBLP:conf/nips/LiuTZYX0YJ024,DBLP:journals/corr/abs-2406-02395} preserve adjacency within a row but push vertical or cross-row neighbors far apart in the sequence. This problem grows with resolution: the sequence length grows quadratically with side length, the recurrent transition has to carry information across longer chains, and positional encodings tuned to one patch grid may no longer represent the same spatial relations at a different scale~\cite{DBLP:conf/cvpr/TianWDHQJ23}. Thus, resolution-scalable vision Mamba requires a serialization rule and a positional model that remain faithful to the 2D image geometry.

\begin{figure*}[!t]
    \centering
    \includegraphics[width=\linewidth]{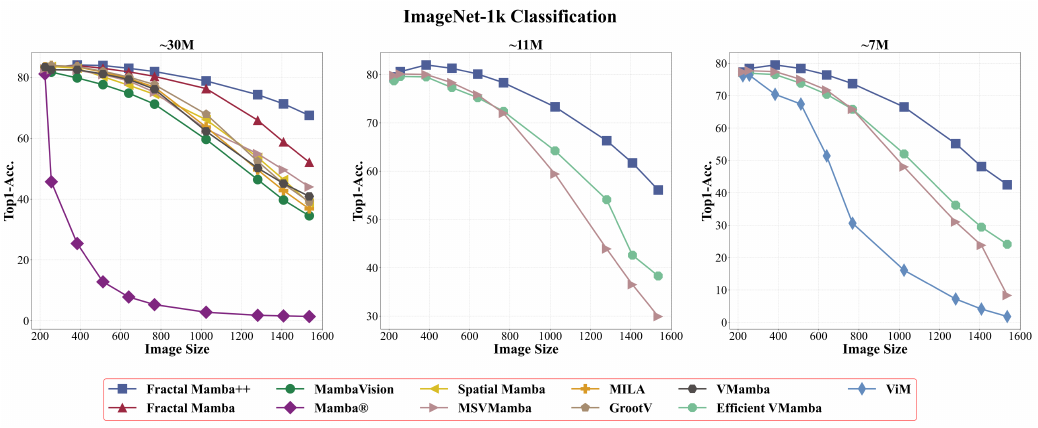}
    \caption{Top-1 classification accuracy of Mamba-based models across different input resolutions on ImageNet-1K. Results are grouped by parameter scale: $\sim$30M, $\sim$11M, and $\sim$7M. Each curve represents model performance from $224^2$ to $1536^2$ input resolution. FractalMamba++ consistently outperforms other methods, especially under large image sizes, demonstrating strong resolution scalability. Comparison methods contain VMamba~\cite{DBLP:conf/nips/LiuTZYX0YJ024}, GrootV~\cite{DBLP:journals/corr/abs-2406-02395}, MILA~\cite{DBLP:conf/nips/HanWXHPGSSZ024}, MSVMamba~\cite{DBLP:conf/nips/ShiDX24}, SpatialMamba~\cite{DBLP:journals/corr/abs-2410-15091}, MambaVision~\cite{DBLP:journals/corr/abs-2407-08083}, Mamba{\textregistered}~\cite{DBLP:journals/corr/abs-2405-14858}, EfficientVMamba~\cite{DBLP:conf/aaai/Pei0X25}, ViM~\cite{DBLP:conf/icml/ZhuL0W0W24} and FractalMamba~\cite{xiao2025boosting}.}
    \label{Introd_Adaptability}
    \vspace{-0.3cm}
\end{figure*}

To address these limitations, we build on the work FractalMamba~\cite{xiao2025boosting} and use the Hilbert curve as the organizing geometry of the backbone. Unlike raster scans whose inter-row jumps scale linearly with grid width, a Hilbert curve recursively traverses a 2D grid while preserving continuity. Spatially adjacent patches are likely to remain close in the 1D sequence, and, because the curve is self-similar, the same local arrangement reappears at every recursion level. This recursive geometry gives a model trained at one resolution consistent neighborhood statistics when evaluated at a different resolution. In FractalMamba++, the Hilbert curve is not only a scan order. It is the common source from which serialization, long-range state shortcuts, and positional encoding are derived.

Fractal serialization alone, however, cannot remove two residual issues that arise when running an SSM over a long 1D sequence. (i)~The recurrent state update chains the transition matrix multiplicatively, so the contribution of earlier patches decays geometrically, and the final hidden state tends to be dominated by later tokens. The top row of Fig.~\ref{Introd_FHSC_Map} visualizes this effect: without FHSC, correlations are biased and localized, indicating weak propagation of information across long sequences. To counter this, we propose a Fractal Hierarchy Skip Connection (FHSC) mechanism that directly exploits the recursion levels of the Hilbert curve. FHSC selects one representative pair of spatially adjacent but sequentially distant sibling segments at each recursion level and adds a gated state injection between their representative positions. Because the skip set is determined purely from the fractal geometry, FHSC is deterministic, resolution-invariant, and can be integrated into the standard SSM computation without custom gradient derivation or dedicated CUDA kernels. The bottom row of Fig.~\ref{Introd_FHSC_Map} shows that this mechanism restores more distributed long-range correlations across the sequence.

\begin{figure*}[!t]
     \centering
     \includegraphics[width=\linewidth]{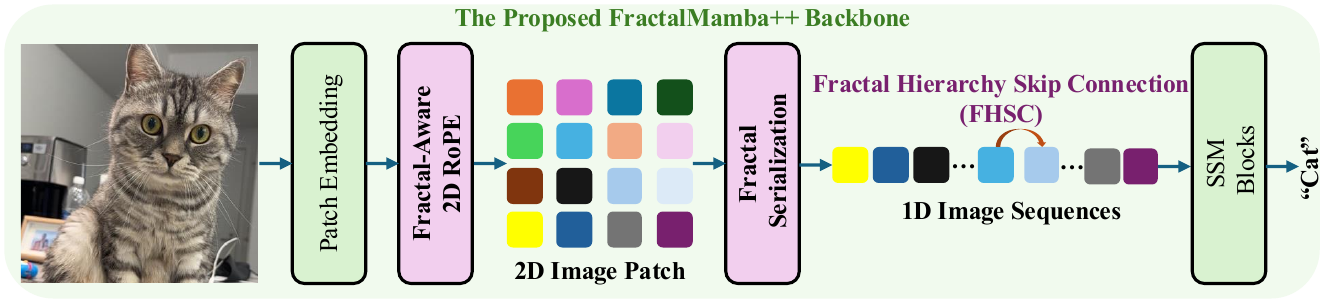}
     \caption{Architecture of the FractalMamba++ backbone. The design contains three Hilbert-geometry-driven components: Fractal-Aware 2D Rotary Position Encoding (FA-RoPE), Fractal Serialization, and Fractal Hierarchy Skip Connection (FHSC).}
     \label{framework}
     \vspace{-0.3cm}
\end{figure*}

(ii)~Even with a locality-preserving scan, the Hilbert curve is not strictly isometric. As illustrated by the red boxes in Fig.~\ref{Introd_FHSC_Map}, some patches that are adjacent in the 2D grid still become distant in the serialized order because they lie on opposite sides of a fractal inflection point. Our previous FractalMamba mitigates such residual locality breaks by introducing multiple offset scan paths, so that corner-adjacent patches are close in at least one path. While effective, this remedy increases path-level complexity and still cannot make a single 1D order preserve every 2D adjacency. FractalMamba++ instead complements Hilbert serialization with Fractal-Aware 2D Rotary Position Encoding (FA-RoPE). FA-RoPE encodes normalized 2D coordinates together with a fractal hierarchy level that describes each patch's structural role in the Hilbert recursion. The resulting positional modulation is resolution-invariant and reflects both actual 2D proximity and recursive fractal geometry rather than only serialized distance. As shown in Fig.~\ref{Introd_Adaptability}, the combination of fractal serialization, FHSC, and FA-RoPE allows FractalMamba++ to adapt to varying resolutions more gracefully than existing vision Mamba backbones. 

Our contributions are summarized as follows:

\begin{itemize}
    \item We serialize image patches with Hilbert-curve-based Fractal Serialization, whose self-similar structure preserves spatial locality across resolutions and gives vision Mamba a faithful 1D representation of the 2D grid.

    \item We propose FHSC, a deterministic long-range state-injection mechanism derived from the recursion levels of the Hilbert curve. By connecting representative fractal sibling segments across recursion levels, FHSC improves information propagation while preserving the geometry-driven and resolution-scalable nature of the backbone.

    \item We design FA-RoPE, a positional encoding that applies rotary transformations over normalized 2D coordinates and a fractal hierarchy level. It replaces scanning-order distance with true 2D proximity and yields resolution-invariant, fractal-aware positional relations.

    \item We validate FractalMamba++ on classification, detection, segmentation, and remote sensing change detection. Extensive results show that the unified design of Fractal Serialization, FHSC, and FA-RoPE improves high-resolution robustness and downstream transferability.
    
\end{itemize}

\section{Related Work}
\subsection{Vision Backbone Architecture}

Vision backbones have been dominated by CNN-based~\cite{DBLP:conf/nips/KrizhevskySH12,DBLP:journals/corr/SimonyanZ14a,DBLP:conf/cvpr/HeZRS16,DBLP:journals/corr/HowardZCKWWAA17,DBLP:conf/cvpr/RadosavovicKGHD20} and ViT-based~\cite{DBLP:conf/nips/VaswaniSPUJGKP17,DBLP:conf/iclr/DosovitskiyB0WZ21,DBLP:conf/iccv/LiuL00W0LG21,DBLP:conf/eccv/TouvronCJ22} models, each carrying distinct inductive biases. CNNs are naturally suited to local connectivity and spatial hierarchies, and classical architectures such as AlexNet, VGG and ResNet have served as performance benchmarks. ViTs introduce a different paradigm by applying self-attention to non-overlapping image patches, which enables the modeling of global context and long-range dependencies and has become a cornerstone of large-scale pretraining frameworks. Despite their representational power, ViTs face two limitations. Their quadratic complexity in sequence length hinders scalability to high-resolution inputs or longer token streams, and they often struggle to generalize across resolutions because their learned positional embeddings are tied to fixed patch grids and do not naturally adapt to scale changes~\cite{DBLP:conf/cvpr/TianWDHQJ23}. These limitations motivate the exploration of backbones that are simultaneously computationally efficient and resolution-flexible. FractalMamba++ contributes to this line of work by combining the efficiency of Mamba with fractal-based spatial structure modeling and a positional encoding that is resolution-invariant, offering a new perspective beyond the CNN and Transformer families.

\subsection{State Space Model}

SSMs have emerged as alternatives to Transformers in sequence modeling, offering linear complexity in sequence length. Representative models such as S4~\cite{DBLP:conf/iclr/GuGR22} and Mamba~\cite{DBLP:conf/icml/DaoG24} replace the explicit attention mechanism with parameterized state transitions, enabling efficient long-range dependency modeling, and Mamba further introduces input-dependent selective updates that improve both expressiveness and scalability. Following their success in language modeling, SSMs have been extended to vision recognition~\cite{DBLP:conf/nips/LiuTZYX0YJ024,DBLP:journals/corr/abs-2406-02395,DBLP:conf/nips/HanWXHPGSSZ024,DBLP:conf/nips/ShiDX24,DBLP:journals/corr/abs-2410-15091,DBLP:journals/corr/abs-2407-08083,DBLP:journals/corr/abs-2405-14858,DBLP:conf/aaai/Pei0X25,DBLP:conf/icml/ZhuL0W0W24,xiao2025boosting}, as well as generative and vision-language tasks~\cite{DBLP:journals/corr/abs-2408-12245,DBLP:conf/aaai/0008ZZDHW25}, by splitting images into non-overlapping patches and serializing them into 1D token streams that the SSM can process.

A central challenge in this adaptation is that visual data is inherently two-dimensional, so naive rasterization methods often break spatial adjacency and distort structural continuity. This issue is especially pronounced under high-resolution inputs, where the sequence length grows quadratically with the image side and any mismatch between 1D order and 2D adjacency is amplified by the recurrent state update. Recent graph-based methods such as GrootV~\cite{DBLP:journals/corr/abs-2406-02395} construct a 4-connected graph over patches and prune it to obtain data-dependent scanning paths. Although flexible, such graph/path construction does not impose the recursive self-similarity required for resolution-invariant neighborhood statistics, and its learned or searched paths are coupled to the training setting. FractalMamba++ follows a different route: it uses a mathematically defined Hilbert curve for traversal, derives FHSC from Hilbert recursion levels rather than from learned graph pruning, and encodes the fractal hierarchy explicitly through FA-RoPE. The overall architecture is shown in Fig.~\ref{framework}.

\section{Method}

In this section we describe FractalMamba++ in detail. Our design is organized around one principle: every component is derived from the recursive self-similar structure of the Hilbert curve, so that the resulting backbone is deterministic, resolution-invariant, and free of data-dependent search. Hilbert-curve-based Fractal Serialization maps the 2D patch grid into a 1D sequence that preserves spatial locality across resolutions (Section~\ref{sec:fractal}). FHSC reuses the recursion levels of the same curve to install a compact set of deterministic long-range state injections that alleviate information fading in the recurrent state (Section~\ref{sec:fhsc}). FA-RoPE turns positional encoding into a function of normalized 2D coordinates and fractal hierarchy level, so that feature interactions reflect true 2D proximity and fractal structure rather than 1D scanning order (Section~\ref{sec:farope}). Together, these three components form a unified framework that improves resolution robustness, spatial structure preservation, and long-range visual understanding, while remaining simple enough to be integrated into SSM-based vision backbones with automatic differentiation.

\subsection{Preliminaries}

\noindent \textbf{Selective State Space Models.}
SSMs provide a principled way to model sequences through a recurrent linear dynamical system. Starting from the continuous-time Linear Time-Invariant (LTI) formulation and discretizing it with a zero-order hold~\cite{DBLP:conf/icml/DaoG24,DBLP:journals/corr/abs-2312-00752}, an SSM can be compactly written as
\begin{equation}
    h_t = \overline{\mathbf{A}}\, h_{t-1} + \overline{\mathbf{B}}\, x_t, \qquad y_t = \mathbf{C}\, h_t,
    \label{first_defi}
\end{equation}
where $h_t$ is the hidden state at step $t$, $x_t$ and $y_t$ are the input and output tokens, and $\overline{\mathbf{A}}, \overline{\mathbf{B}}, \mathbf{C}$ are the discretized state, input and output matrices. A detailed derivation of the discretization and of the convolutional kernel view of Eqn.~\ref{first_defi} is provided in the supplementary material. Classical SSMs share the same $\overline{\mathbf{A}}, \overline{\mathbf{B}}, \mathbf{C}, \Delta$ across all time steps, which limits their expressiveness on heterogeneous inputs. Selective SSMs~\cite{DBLP:journals/corr/abs-2312-00752} address this by making $\overline{\mathbf{B}}, \mathbf{C}$ and the step size $\Delta$ input-dependent, so that the operator becomes a time-varying system that can selectively attend to salient tokens. In FractalMamba++, we adopt Selective SSM as the core sequence operator and design all subsequent modules around the recurrent update rule in Eqn.~\ref{first_defi}.

\subsection{Fractal Serialization}\label{sec:fractal}

\noindent \textbf{Limitations of Linear Curves.}
Selecting appropriate input tokens at each time step is critical for effective representation learning in SSMs, which requires that the serialization of a 2D image into a 1D sequence faithfully retains the image's inherent structural information. Patches that are adjacent in the image should remain close in the 1D token stream, so that the sequence model can exploit this proximity to capture both local and global visual patterns. Existing linear scanning strategies, such as row-major, Z-order and Zigzag curves, only partially satisfy this requirement. They preserve adjacency within individual rows but break inter-row relationships: two patches in vertically adjacent rows are always separated by at least one full row in the sequence, and this gap grows linearly with the grid width. Consequently, the sequential state update in Eqn.~\ref{first_defi} cannot treat vertical neighbors as efficiently as horizontal ones, and the resulting representation is increasingly biased as the input resolution increases.

\noindent \textbf{Fractal Curves.}
We instead adopt the Hilbert curve, a fractal constructed via recursive subdivision of the 2D grid. Its self-similar structure guarantees that spatially adjacent patches stay close in the 1D sequence, and, crucially, that this locality property is preserved at every recursion level. The curve begins at a designated corner, and recursively traverses midpoints along the $\vec{x}$ and $\vec{y}$ directions to visit all patches while maintaining local continuity. A model trained at one resolution encounters the same local neighborhood statistics when evaluated at a different resolution, which provides a robust foundation for resolution-scalable visual modeling and eliminates the linear-in-width inter-row penalty of raster-style scans.

\noindent \textbf{Quantitative Evaluation.}
To quantify the structural fidelity of different scanning strategies, we introduce the Structure Distortion Score (SDS). For a center patch $i$ with 2D coordinate $p_i \in \mathbb{R}^2$ and a $k$-neighborhood $\mathcal{N}_i^{(k)} = \{i \pm 1, \ldots, i \pm k\}$ in the serialized sequence, SDS is defined as
\begin{equation}
    \text{SDS}^{(k)}(i) = \frac{1}{2k}\sum_{j \in \mathcal{N}_i^{(k)}} \| p_i - p_j \|_2,
    \label{eq:sds}
\end{equation}
so that a lower $\text{SDS}^{(k)}$ indicates stronger preservation of local spatial structure within a window of radius $k$. 
For $k=1$, any continuous path that moves to a 4-connected neighbor achieves the minimum, making SDS informative only when $k>1$, where it measures structural preservation beyond immediate adjacency.
We therefore report $\text{SDS}^{(2)}$ in Tab.~\ref{Tab_SDS}.
Under this setting, all patches serialized by the Hilbert curve remain below an SDS of $1.5$, while the linear and local scans exceed this threshold on a substantial portion of the sequence.
This gap confirms the structural advantage of Fractal Serialization over existing linear strategies and is consistent with the resolution-robust behavior observed in our experiments.

\begin{table}[!t]
\centering
\caption{Percentage of patches under an SDS$^{(2)}$ threshold across scanning strategies. Fractal Serialization maintains low structural distortion consistently, while linear scans fail at the $1.5$ threshold. The full table with more thresholds is given in the supplementary material.}
\label{Tab_SDS}
\renewcommand\arraystretch{1.3}
\setlength\tabcolsep{0pt}
\begin{tabular*}{\columnwidth}{@{\extracolsep{\fill}}lcccc@{}}
\toprule[1.2pt]
SDS threshold & $1.3$ & $1.5$ & $1.9$ & $2.5$ \\
\midrule
Linear Serialization  &  0\%    &  3.1\% & 56.2\% & 56.2\% \\
Local Serialization   &  0\%    &  3.1\% & 43.8\% & 93.8\% \\
Fractal Serialization & 62.5\%  & 100\%  & 100\%  & 100\%  \\
\bottomrule[1.2pt]
\end{tabular*}
\end{table}

\subsection{Fractal Hierarchy Skip Connection}
\label{sec:fhsc}

While Fractal Serialization preserves local spatial structure at every recursion level, the resulting sequence is still processed by a recurrent SSM. As the sequence length grows under high-resolution image inputs, the state update in Eqn.~\ref{first_defi} chains the transition matrix $\overline{\mathbf{A}}$ multiplicatively, so the contribution of distant patches decays geometrically and the model loses access to early visual cues. This phenomenon is visible in the first row of Fig.~\ref{Introd_FHSC_Map}, where strong correlations concentrate near the sequence end and earlier patches contribute little to the final hidden state.

A natural remedy is to insert skip connections into the state transition chain~\cite{DBLP:journals/corr/abs-2406-02395}, so that information from distant patches can reach later positions through a shortcut. However, selecting such connections through data-dependent search or graph pruning may introduce construction cost and can tie the routing pattern to the training data or a specific resolution. This is not ideal for our goal of building a resolution-scalable backbone from a fixed geometric principle. We instead observe that the self-similar structure of the Hilbert curve reveals recurring spatial-sequential weak points, which allows us to derive deterministic state injections from the recursion hierarchy.

\noindent \textbf{Skip Connections from Fractal Recursion.}
A Hilbert curve of order $L$ over $N = 4^L$ patches is built by recursively decomposing the grid into quadrants. At recursion level $l \in \{1, \ldots, L\}$, the sequence is partitioned into $4^l$ contiguous segments, each covering a $2^{L-l}\!\times\!2^{L-l}$ spatial block. The U-shaped folding of the curve creates sibling segments that are spatially adjacent in 2D but sequentially separated by intervening siblings, and these positions recur self-similarly across levels. Instantiating all such sibling relations would introduce many extra edges. We therefore use a representative-pair construction. At each level, we choose the canonical sibling group on the Hilbert recursion spine and connect the midpoint of its first child segment to the midpoint of its fourth child segment. Let $\text{mid}(S)$ denote the midpoint index of a segment $S$. The FHSC edge set is
\begin{equation}
    \mathcal{E} = \bigcup_{l=1}^{L} \Big\{\big(\text{mid}(S_l^{(1)}),\, \text{mid}(S_l^{(4)})\big)\Big\},
    \label{eq:fhsc_set}
\end{equation}
where $S_l^{(1)}$ and $S_l^{(4)}$ denote the first and fourth child segments of this canonical level-$l$ sibling group. This definition deliberately creates one representative edge per recursion level, so $|\mathcal{E}| = L = \mathcal{O}(\log N)$. By the self-similarity of the Hilbert curve, the relative positions of these edges depend only on the recursion level and scale proportionally with the grid. Therefore, $\mathcal{E}$ can be precomputed for a given token grid and reused whenever the same relative Hilbert construction is used.

\noindent \textbf{Gated State Injection.}
FHSC injects the selected skip connections into the SSM through a lightweight additive update. For each target index $t$ with predecessor set $\mathcal{P}(t) = \{j \mid (j, t) \in \mathcal{E}\}$, Eqn.~\ref{first_defi} would become
\begin{equation}
    h_t = \overline{\mathbf{A}}_t h_{t-1} + \overline{\mathbf{B}}_t x_t + \gamma \sum_{j \in \mathcal{P}(t)} \overline{\mathbf{A}}_t h_j, \quad y_t = \mathbf{C}_t h_t,
    \label{eq:fhsc_update}
\end{equation}
where $\gamma$ is a learnable scalar gate, shared within a block and initialized to zero so that the augmented recurrence reduces to the vanilla SSM at the start of training. The predecessor index satisfies $j<t$ under the directed Hilbert order, so each injected state $h_j$ has already been produced by the ordinary forward sweep. Since $|\mathcal{P}(t)| \leq 1$ for every $t$ under Eqn.~\ref{eq:fhsc_set}, FHSC adds only $L=\mathcal{O}(\log N)$ gated state injections per block and does not change the asymptotic sequence complexity. Gradients through both the local recurrence and the injected states are handled by standard automatic differentiation.

\noindent \textbf{Relation to Prior Long-Range Routing.}
Prior efforts to add long-range connections to vision Mamba can be grouped into two families. Graph-based methods such as GrootV~\cite{DBLP:journals/corr/abs-2406-02395} construct a 4-connected graph over patches and prune it to learn a scanning path, so their paths are flexible but data-dependent and do not explicitly encode Hilbert self-similarity. Search-based alternatives may keep a fixed scanning curve but select skip connections via a data-dependent criterion, which introduces extra construction complexity and weakens resolution-invariant reuse. FHSC departs from both. It keeps the Hilbert curve as the backbone traversal and adds a small, deterministic set of fractal-level state injections fully determined by the recursive geometry of the curve. This geometry-driven construction removes data-dependent routing, distinguishes the method from generic graph or path pruning, and allows training and deployment with standard SSM operators.

\subsection{Fractal-Aware 2D Rotary Position Encoding}\label{sec:farope}

Even with a locality-preserving scan, the Hilbert curve is not strictly isometric: two patches that are adjacent on the 2D grid may still be pushed apart in the serialized index at fractal inflection points. This misalignment weakens local context modeling, especially when the input resolution changes. Applying a standard Rotary Position Embedding (RoPE)~\cite{DBLP:journals/ijon/SuALPBL24} over the serialized index $n$ does not resolve the issue, because the rotary phase $(n-m)\theta$ still measures distance along the Hilbert order. Two patches separated by a curve inflection therefore remain far apart in the encoded positional relation even if they are neighbors in the image. FA-RoPE resolves this mismatch by encoding the true 2D grid coordinates and a fractal hierarchy level for each patch, rather than the 1D serialized index. A brief review of standard 1D RoPE is provided in the supplementary material.

\begin{table*}[t]
\centering
\setlength\tabcolsep{4.5pt}
\renewcommand\arraystretch{1.4}
\caption{Image classification performance (Top-1 Accuracy) on ImageNet-1K under varying input resolutions. FLOPs are measured at $224 \times 224$. Following MSVMamba~\cite{DBLP:conf/nips/ShiDX24}, we evaluate Tiny (FractalMamba++ (T)), Micro (FractalMamba++ (M)), and Nano (FractalMamba++ (N)) model capacities.}
\label{tab:imagenet_multiscale}
\begin{tabular}{@{}llcc|ccccccccccc@{}}
\toprule[1.5pt]
\multirow{2}{*}{Method} & \multirow{2}{*}{Publication} & \multirow{2}{*}{Param.} & \multirow{2}{*}{FLOPs} &
\multicolumn{10}{c}{Input Resolution} \\
\cmidrule(l){5-14}
 & & & & 224$^2$ & 256$^2$ & 384$^2$ & 512$^2$ & 640$^2$ & 768$^2$ & 1024$^2$ & 1280$^2$ & 1408$^2$ & 1536$^2$ \\
\midrule

VMamba & NeurIPS'24 & 31M & 4.9G &
82.5 & 82.5 & 82.5 & 81.1 & 79.3 & 76.1 & 62.3 & 50.2 & 45.1 & 40.9 \\

GrootV & NeurIPS'24 & 30M & 4.8G &
83.4 & 83.9 & 83.6 & 82.0 & 80.1 & 77.6 & 67.9 & 52.4 & 45.0 & 39.1 \\

MILA & NeurIPS'24 & 25M & 4.2G &
83.5 & 83.9 & 83.5 & 81.7 & 79.6 & 76.8 & 63.7 & 49.6 & 42.8 & 36.8 \\

MSVMamba & NeurIPS'24 & 33M & 4.6G &
82.8 & 82.5 & 82.3 & 80.9 & 78.8 & 75.1 & 63.0 & 54.9 & 49.6 & 44.0 \\

Spatial Mamba & ICLR'25 & 27M & 4.5G &
83.5 & 83.6 & 83.0 & 80.2 & 77.4 & 74.4 & 66.1 & 53.7 & 46.4 & 38.7 \\

Mamba{\textregistered} & CVPR'25 & 29M & 4.6G &
81.1 & 45.7 & 25.4 & 12.8 & 7.8 & 5.3 & 2.8 & 1.8 & 1.6 & 1.4 \\

MambaVision & CVPR'25 & 32M & 4.4G &
82.3 & 81.7 & 79.8 & 77.6 & 74.8 & 71.2 & 59.6 & 46.4 & 39.7 & 34.5 \\
FractalMamba & AAAI'25 & 31M & 4.8G &
83.0 & 83.5 & 83.9 & 83.0 & 81.8 & 80.3 & 76.3 & 65.9 & 58.8 & 52.1 \\

\rowcolor[HTML]{F2F2F2}
FractalMamba++ (T) & Ours & 30M & 4.8G &
83.1 & 83.5 & 84.3 & 83.9 & 83.0 & 81.9 & 79.0 & 74.6 & 71.8 & 67.9 \\

\midrule[1.2pt]

MSVMamba & NeurIPS'24 & 12M & 1.5G &
79.8 & 80.1 & 80.0 & 78.3 & 75.8 & 72.0 & 59.4 & 43.9 & 36.5 & 29.9 \\

Efficient VMamba & AAAI'25 & 11M & 1.3G &
78.7 & 79.6 & 79.5 & 77.3 & 75.2 & 72.4 & 64.2 & 54.1 & 42.6 & 38.3 \\

\rowcolor[HTML]{F2F2F2}
FractalMamba++ (M)& Ours & 11M & 1.5G &
79.6 & 80.6 & 82.2 & 81.4 & 80.1 & 78.5 & 73.6 & 66.7 & 62.3 & 56.7 \\

\midrule[1.2pt]

MSVMamba & NeurIPS'24 & 7M & 0.9G &
77.3 & 77.7 & 77.4 & 75.0 & 71.7 & 65.8 & 48.0 & 31.0 & 23.8 & 18.3 \\

ViM & ICML'24 & 7M & 1.5G &
76.1 & 76.3 & 70.4 & 67.4 & 51.4 & 30.6 & 16.1 & 7.2 & 4.1 & 1.8  \\

Efficient VMamba & AAAI'25 & 6M & 0.8G &
76.5 & 76.9 & 76.5 & 73.8 & 70.4 & 65.8 & 52.0 & 36.2 & 29.4 & 24.1 \\

\rowcolor[HTML]{F2F2F2}
FractalMamba++ (N) & Ours & 7M & 0.9G &
77.4 & 78.4 & 79.7 & 78.5 & 76.4 & 73.9 & 66.8 & 55.8 & 48.7 & 43.0 \\
\bottomrule[1.5pt]
\end{tabular}
\vspace{-0.3cm}
\end{table*}

\noindent \textbf{Position Descriptors.}
For a patch at integer grid position $(x_i, y_i)$ in an $H \times W$ token grid, we first define two resolution-invariant coordinates
\begin{equation}
    \hat{x}_i = x_i / H, \qquad \hat{y}_i = y_i / W,
    \label{eq:farope_norm}
\end{equation}
so that the geometry of the encoding does not change with the input resolution. In addition, we associate each patch with a fractal hierarchy level $\delta_i$, defined as the largest recursion level at which the serialized index $i$ coincides with a segment boundary of the Hilbert curve:
\begin{equation}
    \delta_i = \max\big\{\, l \geq 0 \;\big|\; i \equiv 0 \;\text{mod}\; 4^l \,\big\}, \quad i > 0,
    \label{eq:frac_level}
\end{equation}
with $\delta_0 = 0$ by convention. Intuitively, patches that sit in the interior of a fine sub-curve receive a small $\delta$, while patches that land on major fractal junctions receive a large $\delta$. This scalar therefore exposes the structural role of each patch in the recursion hierarchy, a piece of information that the 2D coordinates alone cannot convey.

\noindent \textbf{Three-Group Rotary Encoding.}
We split the head dimension $d_{\text{head}}$ into three groups of sizes $d_x, d_y, d_\delta$ associated with the three descriptors defined above, with $d_x + d_y + d_\delta = d_{\text{head}}$. Following standard RoPE~\cite{DBLP:journals/ijon/SuALPBL24}, the $t$-th channel of each group uses rotation frequencies
\begin{equation}
    \theta_t^{\star} = b^{-2t/d_\star}, \quad \star \in \{x, y, \delta\},
\end{equation}
with base $b{=}10000$, and every pair of adjacent channels is interpreted as a complex number. In the token-mixing branch where projected feature factors are compared or modulated by positional phase, FA-RoPE applies three independent complex rotations to the patch representation. Using query/key notation only to express this relative-position property, the rotated feature for patch $i$ is
\begin{equation}
    \mathbf{q}_i' = \Big[\mathbf{q}_i^{(x)} \circ e^{\mathrm{i}\hat{x}_i \boldsymbol{\theta}^x},\; \mathbf{q}_i^{(y)} \circ e^{\mathrm{i}\hat{y}_i \boldsymbol{\theta}^y},\; \mathbf{q}_i^{(\delta)} \circ e^{\mathrm{i}\delta_i \boldsymbol{\theta}^\delta}\Big],
    \label{eq:farope_q}
\end{equation}
and the same operation is applied to the paired feature $\mathbf{k}_j$ to obtain $\mathbf{k}_j'$. The resulting position-dependent interaction score
\begin{equation}
    \mathbf{A}_{ij}' = \operatorname{Re}\!\left[\mathbf{q}_i'\,(\mathbf{k}_j')^{*}\right]
\end{equation}
depends only on the relative descriptors $(\hat{x}_i-\hat{x}_j,\; \hat{y}_i-\hat{y}_j,\; \delta_i-\delta_j)$. Thus, patches that are spatially close receive compatible positional phases even when their serialized indices are far apart. The notation follows RoPE but the role of FA-RoPE in FractalMamba++ is broader than self-attention: it supplies a resolution-invariant, structure-aware positional modulation for token interactions inside the SSM-based backbone.

\noindent \textbf{Why FA-RoPE Complements Fractal Serialization.}
Three properties make FA-RoPE a natural companion to fractal serialization. First, the rotary phase depends on $(\hat{x}, \hat{y})$ rather than on the serialized index, so positional relations are no longer distorted at fractal inflection points. Second, the normalization by $(H, W)$ makes the encoding resolution-invariant, allowing a model trained at one resolution to be evaluated at a larger resolution without re-interpolating positional parameters. Third, the fractal hierarchy level $\delta$ supplies a structural signal that is specific to the Hilbert traversal and not recoverable from the 2D coordinates alone, so the model can distinguish major fractal junctions from positions inside smooth sub-curves.

\section{Experiments}
We evaluate FractalMamba++ against several established Mamba-based baselines across four vision tasks: image classification, object detection and instance segmentation, semantic segmentation, and remote sensing binary change detection. These tasks span resolutions from $224^2$ to $1536^2$ and provide a comprehensive testbed for the multi-scale adaptability of our model. Particular attention is paid to the remote sensing change detection task, which operates at a native resolution of $1024 \times 1024$ and places strong demands on long-range spatial consistency. All experiments are conducted on 8 NVIDIA H800 GPUs. Following MSVMamba~\cite{DBLP:conf/nips/ShiDX24}, we evaluate three model capacities, denoted Tiny (T), Micro (M) and Nano (N).

\begin{table}[t]
\centering
\caption{Inference speed comparison of Mamba-based models on ImageNet-1K at $224 \times 224$ input.}
\label{tab:inference_speed}
\renewcommand\arraystretch{1.3}
\setlength\tabcolsep{0pt}
\begin{tabular*}{\columnwidth}{@{\extracolsep{\fill}}lcc@{}}
\toprule[1.5pt]
Method & Parameters & FPS \\
\midrule
VMamba         & 31M & 1916  \\
GrootV        & 30M & 315  \\
MILA           & 25M & 947  \\
MSVMamba       & 33M & 684  \\
Spatial Mamba  & 27M & 957  \\
Mamba{\textregistered} & 29M & 1044  \\
MambaVision    & 32M & 3426  \\
\rowcolor[HTML]{F2F2F2}
FractalMamba++ (T) & 30M & 361  \\
\midrule
MSVMamba       & 12M & 1000  \\
Efficient VMamba & 11M & 2187  \\
\rowcolor[HTML]{F2F2F2}
FractalMamba++ (M) & 11M & 572 \\
\midrule
MSVMamba       & 7M & 1181  \\
ViM            & 7M & 398  \\
Efficient VMamba & 6M & 2380  \\
\rowcolor[HTML]{F2F2F2}
FractalMamba++ (N) & 7M & 682  \\
\bottomrule[1.5pt]
\end{tabular*}
\vspace{-0.5cm}
\end{table}

\subsection{Image Classification}

We train FractalMamba++ from scratch on ImageNet-1K~\cite{deng2009imagenet} for 300 epochs with 20-epoch warm-up and  batch size of 1024. Optimization uses AdamW~\cite{loshchilov2017decoupled} with $\beta_1{=}0.9$, $\beta_2{=}0.999$ and  weight decay of $0.05$, following cosine learning rate schedule starting at $10^{-3}$. We further apply label smoothing of $0.1$ and exponential moving average (EMA) to stabilize training. Models are evaluated across a broad range of resolutions, from $224^2$ to $1536^2$. At 30M scale, we compare against VMamba~\cite{DBLP:conf/nips/LiuTZYX0YJ024}, GrootV~\cite{DBLP:journals/corr/abs-2406-02395}, MILA~\cite{DBLP:conf/nips/HanWXHPGSSZ024}, MSVMamba~\cite{DBLP:conf/nips/ShiDX24}, SpatialMamba~\cite{DBLP:journals/corr/abs-2410-15091}, MambaVision~\cite{DBLP:journals/corr/abs-2407-08083}, Mamba{\textregistered}~\cite{DBLP:journals/corr/abs-2405-14858}, and FractalMamba~\cite{xiao2025boosting}. At 11M and 7M scales, we compare against Efficient VMamba~\cite{DBLP:conf/aaai/Pei0X25}, MSVMamba and ViM~\cite{DBLP:conf/icml/ZhuL0W0W24}. All models are evaluated with their official checkpoints under a unified testing pipeline for fairness.

The detailed accuracy comparisons are reported in Tab.~\ref{tab:imagenet_multiscale}. This multi-resolution classification benchmark is the primary evidence for resolution scalability. At low resolutions such as $224^2$ and $384^2$, FractalMamba++ performs on par with or slightly better than strong Mamba-based models, indicating that the geometry-driven components do not compromise the standard training regime. The advantage becomes much more pronounced as the input size grows. At $1024^2$, the 30M variant reaches 79.0 top-1 accuracy, while SpatialMamba, GrootV and VMamba drop to 66.1, 67.9 and 62.3, respectively. The gap continues to widen at $1280^2$ and beyond, where FractalMamba++ (T) maintains 74.6 accuracy, clearly outperforming FractalMamba, MSVMamba and other Mamba-based baselines. Similar trends are observed at the 11M and 7M scales: FractalMamba++ (M) achieves 73.6 at $1024^2$, and FractalMamba++ (N) reaches 66.8 at the same resolution, both substantially outperforming comparable-capacity baselines. Overall, this behavior supports the central claim that Fractal Serialization, FHSC and FA-RoPE jointly preserve locality and global context under high-resolution inputs, where standard vision Mamba models degrade most sharply.

To complement the accuracy evaluation, we measure inference efficiency on ImageNet-1K at a $224 \times 224$ resolution, with batch size 32 on a H800 GPU. As summarized in Tab.~\ref{tab:inference_speed}, FractalMamba++ runs at 361/572/682 FPS for the T/M/N variants. It is not the fastest model in the comparison; VMamba, Efficient VMamba and MambaVision remain faster at standard resolution. At the 30M scale, however, FractalMamba++ is comparable to GrootV in speed while providing much stronger high-resolution accuracy. The FHSC and FA-RoPE operations introduce moderate overhead, while FHSC itself adds only $\mathcal{O}(\log N)$ gated state injections per block and can be implemented with standard automatic differentiation. The key trade-off is therefore accuracy under high-resolution scaling versus standard-resolution latency: FractalMamba++ offers a favorable accuracy-efficiency balance in the high-resolution regime where several faster baselines lose substantial accuracy.

\begin{table}[t]
\caption{Object detection and instance segmentation performance on COCO using Mask R-CNN. FLOPs are measured at input resolution of $1280 \times 800$.}
\label{tab:coco_detection}
\renewcommand\arraystretch{1.3}
\setlength\tabcolsep{0pt}
\centering
\scriptsize
\begin{tabular*}{\columnwidth}{@{\extracolsep{\fill}}lcccccc@{}}
\toprule[1.5pt]
\multirow{2}{*}{Method} 
& \multirow{2}{*}{Param.} 
& \multirow{2}{*}{FLOPs} 
& \multicolumn{2}{c}{1$\times$ Schedule} 
& \multicolumn{2}{c}{3$\times$ Schedule} \\
\cmidrule(lr){4-5} \cmidrule(l){6-7}
& & 
& $AP^{\text{box}}$ 
& $AP^{\text{mask}}$ 
& $AP^{\text{box}}$ 
& $AP^{\text{mask}}$ \\
\midrule
VMamba             & 42M & 286G & 46.5 & 42.1 & 48.5 & 43.2 \\
GrootV            & *   & 265G & 47.0 & 42.7 & 49.0 & 43.8 \\
MSVMamba           & 53M & 252G & 46.9 & 42.2 & 48.3 & 43.2 \\
SpatialMamba       & 46M & 261G & 47.6 & 42.9 & 49.3 & 43.6 \\
DefMamba           & *   & 268G & 47.5 & 42.8 & *    & *    \\
FractalMamba       & 41M & 266G & 46.8 & 42.4 & 48.5 & 43.3 \\
\rowcolor[HTML]{F2F2F2}
FractalMamba++ (T) & 41M & 258G & 48.1 & 43.4 & 49.8 & 44.5 \\
\midrule[1.2pt]
MSVMamba           & 32M & 201G & 43.8 & 39.9 & 46.3 & 41.8 \\
Efficient VMamba   & 31M & 197G & 39.3 & 36.7 & 41.6 & 38.2 \\
\rowcolor[HTML]{F2F2F2}
FractalMamba++ (M) & 31M & 198G & 44.4 & 40.5 & 47.2 & 42.5 \\
\bottomrule[1.5pt]
\end{tabular*}
\vspace{-0.5cm}
\end{table}

\subsection{Object Detection}

We evaluate FractalMamba++ on object detection and instance segmentation using the MS COCO 2017 dataset~\cite{lin2014microsoft}, with Mask R-CNN~\cite{DBLP:conf/iccv/HeGDG17} as the detector under MMDetection~\cite{chen2019mmdetection} and following the training setup of Swin-T~\cite{DBLP:conf/iccv/LiuL00W0LG21}. All backbones are initialized from ImageNet-1K pre-trained weights and fine-tuned with AdamW~\cite{loshchilov2017decoupled} at an initial learning rate of $1 \times 10^{-4}$, under both the 1$\times$ (12-epoch) and 3$\times$ (36-epoch) schedules. The input resolution is fixed at $1280 \times 800$ with a batch size of 16, and we apply multi-scale training and random horizontal flipping as is standard.

Tab.~\ref{tab:coco_detection} compares FractalMamba++ with VMamba~\cite{DBLP:conf/nips/LiuTZYX0YJ024}, GrootV~\cite{DBLP:journals/corr/abs-2406-02395}, MSVMamba~\cite{DBLP:conf/nips/ShiDX24}, Efficient VMamba~\cite{DBLP:conf/aaai/Pei0X25}, SpatialMamba~\cite{DBLP:journals/corr/abs-2410-15091}, DefMamba~\cite{liu2025defmamba} and FractalMamba~\cite{xiao2025boosting}. Under the 1$\times$ schedule, the 41M variant obtains 48.1 mAP for box prediction and 43.4 mAP for mask prediction, outperforming SpatialMamba (47.6/42.9), GrootV (47.0/42.7), MSVMamba (46.9/42.2), and VMamba (46.5/42.1). Under the longer 3$\times$ schedule, FractalMamba++ further improves to 49.8 and 44.5 for box and mask mAP, which are the best results among all compared models. A similar trend is observed at the 31M scale, where FractalMamba++ achieves 44.4/40.5 under the 1$\times$ schedule and 47.2/42.5 under the 3$\times$ schedule, consistently outperforming comparable-capacity baselines. These results show that the locality preservation and long-range context modeling brought by Fractal Serialization, FHSC and FA-RoPE transfer to dense prediction. We do not use COCO as evidence of extreme-resolution robustness; rather, it validates that the learned representation remains useful for object-level localization and mask prediction.

\begin{table}[t]
\centering
\caption{Semantic segmentation results on ADE20K using UPerNet with input resolution $512 \times 2048$. FLOPs are computed under the same setting. SS and MS denote single-scale and multi-scale inference.}
\label{tab:ade20k_segmentation}
\renewcommand\arraystretch{1.4}
\setlength\tabcolsep{0pt}
\begin{tabular*}{\columnwidth}{@{\extracolsep{\fill}}lcccc@{}}
\toprule[1.5pt]
Method & Param. & FLOPs & mIoU (SS) & mIoU (MS) \\
\midrule

VMamba & 55M & 946G & 47.3 & 48.3 \\
GrootV & * & 941G & 48.5 & 49.4 \\
MSVMamba & 65M & 942G & 47.6 & 48.5 \\
Spatial Mamba & 57M & 936G & 48.6 & 49.4 \\
DefMamba & 65M & 946G & 48.7 & 49.6 \\
MambaVision & 55M & 945G & 46.0 & * \\
Mamba{\textregistered} & 56M & * & 45.3 & * \\
FractalMamba & 53M & 942G & 48.0 & 48.9 \\

\rowcolor[HTML]{F2F2F2}
FractalMamba++ (T) & 62M & 956G & 49.3 & 50.1 \\

\midrule[1.2pt]

MSVMamba & 42M & 875G & 45.1 & 45.4 \\
ViM & 46M & * & 44.9 & * \\

\rowcolor[HTML]{F2F2F2}
FractalMamba++ (M) & 44M & 887G & 46.1 & 46.6 \\

\midrule[1.2pt]

Efficient VMamba & 29M & 505G & 41.5 & 42.1 \\
Local-ViM & 36M & 181G & 43.4 & 44.4 \\
PlainMamba & 35M & 174G & 44.1 & 44.6 \\

\rowcolor[HTML]{F2F2F2}
FractalMamba++ (N) & 37M & 858G & 44.9 & 45.5 \\
\bottomrule[1.5pt]
\end{tabular*}
\vspace{-0.5cm}
\end{table}

\subsection{Semantic Segmentation}

We further assess the segmentation capability of FractalMamba++ on the ADE20K dataset~\cite{zhou2019semantic} using UPerNet~\cite{xiao2018unified} as the segmentation framework, following the training protocol of Swin~\cite{DBLP:conf/iccv/LiuL00W0LG21}. The backbone is initialized with ImageNet-1K pre-trained weights and fine-tuned for 160K iterations with AdamW at a learning rate of $6 \times 10^{-5}$ and a batch size of 16. The training crop size is $512 \times 512$, while FLOPs are measured at $512 \times 2048$ for consistency with prior work. We report both single-scale (SS) and multi-scale (MS) mIoU.

Tab.~\ref{tab:ade20k_segmentation} summarizes the results across three model capacities. At the tiny scale (62M parameters), FractalMamba++ achieves 49.3 mIoU (SS) and 50.1 mIoU (MS), outperforming DefMamba (48.7/49.6), SpatialMamba (48.6/49.4), GrootV (48.5/49.4), MSVMamba (47.6/48.5), and FractalMamba (48.0/48.9). At the micro scale (44M), it obtains 46.1/46.6, exceeding MSVMamba-M (45.1/45.4). At the nano scale (37M), it reaches 44.9/45.5, surpassing Local-ViM (43.4/44.4) and PlainMamba (44.1/44.6). Since semantic segmentation requires spatially coherent representations over dense grids, these consistent gains indicate that Fractal Serialization, FHSC and FA-RoPE improve structure-aware representation beyond image classification.

\begin{table}[t]
\centering
\caption{Performance and efficiency comparison on binary change detection using the LEVIR-CD+ dataset.}
\label{tab:cd_levir}
\renewcommand\arraystretch{1.4}
\setlength\tabcolsep{4pt}
\begin{tabular}{@{}lcc|ccccc@{}}
\toprule[1.5pt]
Method & Param. & FLOPs & IoU & Precision & Recall & KC & F1 \\
\midrule
ChangeMamba & 17M & 46G & 78.6 & 88.8 & 87.3 & 87.5 & 88.0 \\
ChangeMamba-S & 50M & 115G & 78.3 & 89.2 & 86.5 & 87.3 & 87.8 \\
ChangeMamba-B & 85M & 179G & 79.2 & 89.2 & 87.6 & 87.9 & 88.4 \\
FractalMamba & 35M & 55G & 80.0 & 89.3 & 88.4 & 88.4 & 89.9 \\
\rowcolor[HTML]{F2F2F2}
FractalMamba++ & 35M & 56G & {80.8} & {90.3} & {88.7} & {89.1} & {90.4} \\
\bottomrule[1.5pt]
\end{tabular}
\vspace{-0.5cm}
\end{table}

\subsection{Remote Sensing Binary Change Detection}

We additionally evaluate FractalMamba++ on the binary change detection (BCD) task using the LEVIR-CD+ dataset, which is an enhanced version of LEVIR-CD and contains 985 pairs of very high-resolution aerial images at $1024 \times 1024$ with a $0.5$-meter spatial resolution. Following the protocol of ChangeMamba~\cite{chen2404changemamba}, we adopt the same Change-Decoder, initialize the backbone from ImageNet pre-trained weights, and train with AdamW at a learning rate of $10^{-4}$, weight decay $0.005$ and a batch size of 16. The model is evaluated with five widely adopted metrics: Intersection over Union (IoU), Precision, Recall, Kappa Coefficient (KC) and F1 Score.

As reported in Tab.~\ref{tab:cd_levir}, FractalMamba++ achieves the best performance across all five metrics. It attains an IoU of 80.8 and an F1 Score of 90.4, improving over FractalMamba (80.0/89.9) with the same parameter scale and a comparable FLOP budget (35M, 56G vs. 35M, 55G). Compared with ChangeMamba-B, which uses 85M parameters and 179G FLOPs, FractalMamba++ delivers higher accuracy with less than one third of the computational cost. Because LEVIR-CD+ uses paired $1024 \times 1024$ remote sensing images, this task provides strong downstream evidence that the proposed geometry-driven design preserves high-resolution spatial consistency in practical dense prediction.

\subsection{Ablation Studies} \label{Analysis}
To understand how each component contributes to the performance, we conduct a series of ablation studies on ImageNet-1K. Unless otherwise specified, all experiments use the Tiny configuration of FractalMamba++ and are evaluated at representative resolutions, covering both the standard training and high-resolution regimes that our design targets.

\begin{table}[t]
\centering
\caption{Ablation and generality analysis on ImageNet-1K. The upper block incrementally builds FractalMamba++ on VMamba, while the lower block applies the proposed Fractal Serialization, FHSC and FA-RoPE to ViM to evaluate generality across Mamba backbones.}
\label{tab:ablation_generality}
\renewcommand\arraystretch{1.3}
\setlength\tabcolsep{0pt}
\scriptsize
\begin{tabular*}{\columnwidth}{@{\extracolsep{\fill}}lccccccc@{}}
\toprule[1.5pt]
Backbone & Fractal & FHSC & FA-RoPE & 224$^2$ & 384$^2$ & 640$^2$ & 1024$^2$ \\
\midrule
\multicolumn{8}{@{}l}{\textit{Incremental ablation on VMamba}} \\
VMamba & \ding{55} & \ding{55} & \ding{55} & 82.5 & 82.5 & 79.3 & 62.3 \\
VMamba & \ding{51} & \ding{55} & \ding{55} & 82.7 & 82.8 & 80.6 & 72.3 \\
VMamba & \ding{51} & \ding{51} & \ding{55} & 83.1 & 83.7 & 82.2 & 76.9 \\
VMamba & \ding{51} & \ding{51} & \ding{51} & 83.1 & 84.3 & 83.0 & 79.0 \\
\midrule[1.2pt]
\multicolumn{8}{@{}l}{\textit{Generality on ViM}} \\
ViM    & \ding{55} & \ding{55} & \ding{55} & 76.1 & 70.4 & 51.4 & 16.1 \\
ViM    & \ding{51} & \ding{55} & \ding{55} & 77.9 & 75.2 & 62.3 & 39.7 \\
ViM    & \ding{51} & \ding{51} & \ding{55} & 78.3 & 75.9 & 64.1 & 42.1 \\
ViM    & \ding{51} & \ding{51} & \ding{51} & 78.5 & 76.2 & 64.7 & 43.8 \\
\bottomrule[1.5pt]
\end{tabular*}
\vspace{-0.3cm}
\end{table}

\begin{table}[t]
\centering
\caption{The upper block compares different fractal curves in FractalMamba++, while the lower block compares Fractal Serialization with multi-scale training using VMamba as the backbone. For MS-Train, resolutions are randomly sampled from \{224, 384, 512\} during training.}
\label{tab:additional_ablation}
\renewcommand\arraystretch{1.3}
\setlength\tabcolsep{0pt}
\begin{tabular*}{\columnwidth}{@{\extracolsep{\fill}}llcccc@{}}
\toprule[1.5pt]
Setting & Method & 224$^2$ & 384$^2$ & 640$^2$ & 1024$^2$ \\
\midrule
\multirow{3}{*}{Curve Type}
& Hilbert Curve  & 83.1 & 84.3 & 83.0 & 79.0 \\
& Coil Curve     & 82.9 & 84.4 & 82.8 & 78.5 \\
& Meurthe Curve  & 83.1 & 84.0 & 82.6 & 78.6 \\
\midrule[1.2pt]
\multirow{2}{*}{Training Strategy}
& MS-Train       & 82.7 & 82.8 & 79.5 & 62.4 \\
& Fractal Only   & 82.7 & 82.8 & 80.6 & 72.3 \\
\bottomrule[1.5pt]
\end{tabular*}
\vspace{-0.5cm}
\end{table}

\subsubsection{Effectiveness and Generality of Proposed Modules}
Tab.~\ref{tab:ablation_generality} evaluates the proposed modules on ImageNet-1K. The upper block starts from VMamba and progressively adds Fractal Serialization, FHSC, and FA-RoPE. Fractal Serialization raises the $1024^2$ accuracy from $62.3$ to $72.3$, confirming the importance of locality-preserving traversal for resolution scalability. FHSC further improves it to $76.9$ by adding deterministic fractal-level state routes, and FA-RoPE pushes the result to $79.0$ by reinforcing 2D and fractal-aware positional cues. These complementary gains support our unified Hilbert-geometry design.
The lower block applies the same components to ViM. Fractal Serialization improves the $1024^2$ accuracy from $16.1$ to $39.7$, FHSC raises it to $42.1$, and FA-RoPE further improves it to $43.8$. This trend shows that the proposed modules are not tied to a backbone implementation, and that serialization, state routing, and positional encoding work best when derived from the same Hilbert geometry.

\subsubsection{Fractal Traversal Design versus Multi-scale Training}
Tab.~\ref{tab:additional_ablation} summarizes two ablations on ImageNet-1K. The upper block compares Hilbert with Coil and Meurthe curves in FractalMamba++. All three curves perform competitively, indicating that fractal traversal provides a robust structural prior. Hilbert gives the strongest high-resolution results, reaching 83.0 at $640^2$ and 79.0 at $1024^2$, so we use it as the default traversal.
The lower block compares Fractal Serialization with multi-scale training using VMamba as the backbone. MS-Train samples resolutions from $\{224, 384, 512\}$ during training, while Fractal Only trains at $224^2$ and changes only the serialization rule. Although both methods are similar at low resolutions, Fractal Only achieves 80.6 at $640^2$ and 72.3 at $1024^2$, outperforming MS-Train by 1.1 and 9.9 points. This suggests that preserving geometric structure during serialization is more effective for high-resolution generalization than exposure to multiple training resolutions alone.

\begin{table}[t]
\centering
\caption{Design ablations of FHSC and FA-RoPE on ImageNet-1K. The upper block evaluates different long-range state-injection rules under the same Fractal Serialization backbone. The lower block evaluates different positional encoding variants after Fractal Serialization and FHSC are enabled.}
\label{tab:fhsc_farope_design}
\renewcommand\arraystretch{1.3}
\setlength\tabcolsep{0pt}
\scriptsize
\begin{tabular*}{\columnwidth}{@{\extracolsep{\fill}}llcccc@{}}
\toprule[1.5pt]
Module & Variant & Cost & 384$^2$ & 640$^2$ & 1024$^2$ \\
\midrule
\multicolumn{6}{@{}l}{\textit{FHSC edge design}} \\
FHSC & No FHSC & -- & 82.8 & 80.6 & 72.3 \\
FHSC & Random edges & $\mathcal{O}(\log N)$ & 82.9 & 81.0 & 72.5 \\
FHSC & Uniform interval edges & $\mathcal{O}(\log N)$ & 83.2 & 81.4 & 73.2 \\
FHSC & Hilbert-recursive edges & $\mathcal{O}(\log N)$ & 83.7 & 82.2 & 76.9 \\
\midrule[1.2pt]
\multicolumn{6}{@{}l}{\textit{FA-RoPE design}} \\
FA-RoPE & No FA-RoPE & -- & 83.7 & 82.2 & 76.9 \\
FA-RoPE & Serialized 1D RoPE & -- & 83.8 & 82.5 & 77.4 \\
FA-RoPE & Normalized 2D RoPE & -- & 84.0 & 82.8 & 78.7 \\
FA-RoPE & Normalized 2D RoPE + $\delta$ & -- & 84.3 & 83.0 & 79.0 \\
\bottomrule[1.5pt]
\end{tabular*}
\vspace{-0.5cm}
\end{table}

\subsubsection{Design Analysis of FHSC and FA-RoPE}
Tab.~\ref{tab:fhsc_farope_design} analyzes the design choices of FHSC and FA-RoPE. In the upper block, all FHSC variants use the same Fractal Serialization backbone and the same $\mathcal{O}(\log N)$ edge budget. The no-FHSC baseline reaches 72.3 at $1024^2$. Random edges bring a marginal gain to 72.5, and uniformly spaced edges improve the result to 73.2. In contrast, the proposed Hilbert-recursive edges achieve 76.9, showing that the improvement does not simply come from adding long-range shortcuts. Instead, placing state injections according to the recursive geometry of the Hilbert curve is crucial for effective long-range propagation.

The lower block studies the positional encoding design after Fractal Serialization and FHSC are enabled. Without FA-RoPE, the model obtains 76.9 at $1024^2$. Serialized 1D RoPE improves the result slightly to 77.4, while normalized 2D RoPE raises it to 78.7 by modeling true spatial proximity instead of serialized distance. Adding the fractal hierarchy level $\delta$ gives the best performance, reaching 84.3 at $384^2$, 83.0 at $640^2$, and 79.0 at $1024^2$. This confirms that $\delta$ provides useful Hilbert-structural information beyond 2D coordinates alone, supporting the fractal-aware design of FA-RoPE.

\subsubsection{Generalization to Downstream Tasks}

To further assess the generalization capability of FractalMamba++, we conduct experiments on high-resolution salient object detection (HRSOD). We use PGNet~\cite{DBLP:conf/cvpr/XieXMZC022} as the base framework and replace its Swin-T backbone with FractalMamba++. Performance is evaluated using maxF and BDE.
As shown in Tab.~\ref{tab:generalization_tasks}, FractalMamba++ improves the maxF score from 0.931 to 0.950 and reduces BDE from 46.92 to 44.72 compared with the Swin-T backbone, validating its effectiveness on the downstream high-resolution dense prediction task.

\begin{table}[t]
\centering
\caption{Generalization performance on the downstream HRSOD task. We replace the backbone of PGNet and report maxF and BDE.}
\label{tab:generalization_tasks}
\renewcommand\arraystretch{1.3}
\setlength\tabcolsep{0pt}
\begin{tabular*}{\columnwidth}{@{\extracolsep{\fill}}lccc@{}}
\toprule[1.5pt]
Framework & Backbone & maxF$\uparrow$ & BDE$\downarrow$ \\
\midrule
PGNet & Swin-T & 0.931 & 46.92 \\
\rowcolor[HTML]{F2F2F2}
PGNet & FractalMamba++ & \textbf{0.950} & \textbf{44.72} \\
\bottomrule[1.5pt]
\end{tabular*}
\vspace{-0.5cm}
\end{table}

\section{Conclusion}
We presented FractalMamba++, a resolution-scalable vision backbone built on a unified Hilbert-geometry design. Hilbert-curve-based Fractal Serialization preserves 2D locality during 2D-to-1D conversion and maintains consistent neighborhood structure across resolutions. FHSC reuses the same recursion levels to derive compact deterministic state-injection routes, mitigating long-sequence information fading without data-dependent routing or custom gradient design. FA-RoPE models positional relations through normalized 2D coordinates and a fractal hierarchy level, so token interactions reflect spatial proximity and recursive structure rather than serialized distance. Experiments on ImageNet-1K, COCO, ADE20K and LEVIR-CD+ show that the three components provide complementary gains, especially under high-resolution inputs. These results suggest that scan geometry can serve not only as an ordering rule, but also as a principle for state propagation and positional modeling in sequence-based vision architectures.

{\small
\bibliographystyle{IEEEtran}
\bibliography{IEEEfull}

@inproceedings{DBLP:conf/iccv/HeGDG17,
  author       = {Kaiming He and
                  Georgia Gkioxari and
                  Piotr Doll{\'{a}}r and
                  Ross B. Girshick},
  title        = {Mask {R-CNN}},
  booktitle    = {{ICCV}},
  pages        = {2980--2988},
  publisher    = {{IEEE} Computer Society},
  year         = {2017}
}

@article{ravi2024sam2,
  title={SAM 2: Segment Anything in Images and Videos},
  author={Ravi, Nikhila and Gabeur, Valentin and Hu, Yuan-Ting and Hu, Ronghang and Ryali, Chaitanya and Ma, Tengyu and Khedr, Haitham and R{\"a}dle, Roman and Rolland, Chloe and Gustafson, Laura and Mintun, Eric and Pan, Junting and Alwala, Kalyan Vasudev and Carion, Nicolas and Wu, Chao-Yuan and Girshick, Ross and Doll{\'a}r, Piotr and Feichtenhofer, Christoph},
  journal={arXiv preprint arXiv:2408.00714},
  url={https://arxiv.org/abs/2408.00714},
  year={2024}
}

@article{chen2024far,
  title={How Far Are We to GPT-4V? Closing the Gap to Commercial Multimodal Models with Open-Source Suites},
  author={Chen, Zhe and Wang, Weiyun and Tian, Hao and Ye, Shenglong and Gao, Zhangwei and Cui, Erfei and Tong, Wenwen and Hu, Kongzhi and Luo, Jiapeng and Ma, Zheng and others},
  journal={arXiv preprint arXiv:2404.16821},
  year={2024}
}

@String(CVPR= {IEEE Conf. Comput. Vis. Pattern Recog.})

@String(ICCV= {Int. Conf. Comput. Vis.})

@String(ECCV= {Eur. Conf. Comput. Vis.})

@String(NIPS= {Adv. Neural Inform. Process. Syst.})

@String(ICLR = {Int. Conf. Learn. Represent.})

@String(AAAI = {AAAI})

@String(CVPR  = {CVPR})

@String(ICCV  = {ICCV})

@String(ECCV  = {ECCV})

@String(NIPS  = {NeurIPS})

@String(ICLR  = {ICLR})

@inproceedings{DBLP:conf/naacl/DevlinCLT19,
  author       = {Jacob Devlin and
                  Ming{-}Wei Chang and
                  Kenton Lee and
                  Kristina Toutanova},
  title        = {{BERT:} Pre-training of Deep Bidirectional Transformers for Language
                  Understanding},
  booktitle    = {{NAACL}},
  pages        = {4171--4186},
  year         = {2019}
}

@inproceedings{DBLP:conf/cvpr/TianWDHQJ23,
  author       = {Rui Tian and
                  Zuxuan Wu and
                  Qi Dai and
                  Han Hu and
                  Yu Qiao and
                  Yu{-}Gang Jiang},
  title        = {ResFormer: Scaling ViTs with Multi-Resolution Training},
  booktitle    = {{CVPR}},
  pages        = {22721--22731},
  publisher    = {{IEEE}},
  year         = {2023}
}

@inproceedings{xiao2025boosting,
  title={Boosting Vision State Space Model with Fractal Scanning},
  author={Xiao, Haoke and Tang, Lv and Jiang, Peng-tao and Zhang, Hao and Chen, Jinwei and Li, Bo},
  booktitle={Proceedings of the AAAI Conference on Artificial Intelligence},
  volume={39},
  number={8},
  pages={8646--8654},
  year={2025}
}

@inproceedings{DBLP:conf/cvpr/XieXMZC022,
  author       = {Chenxi Xie and
                  Changqun Xia and
                  Mingcan Ma and
                  Zhirui Zhao and
                  Xiaowu Chen and
                  Jia Li},
  title        = {Pyramid Grafting Network for One-Stage High Resolution Saliency Detection},
  booktitle    = {{CVPR}},
  pages        = {11707--11716},
  publisher    = {{IEEE}},
  year         = {2022}
}

@inproceedings{DBLP:conf/cvpr/RadosavovicKGHD20,
  author       = {Ilija Radosavovic and
                  Raj Prateek Kosaraju and
                  Ross B. Girshick and
                  Kaiming He and
                  Piotr Doll{\'{a}}r},
  title        = {Designing Network Design Spaces},
  booktitle    = {{CVPR}},
  pages        = {10425--10433},
  publisher    = {{IEEE}},
  year         = {2020}
}

@article{DBLP:journals/jmlr/ChowdheryNDBMRBCSGSSTMRBTSPRDHPBAI23,
  author       = {Aakanksha Chowdhery and
                  Sharan Narang and
                  Jacob Devlin and
                  Maarten Bosma and
                  Gaurav Mishra and
                  Adam Roberts and
                  Paul Barham and
                  Hyung Won Chung and
                  Charles Sutton and
                  Sebastian Gehrmann and
                  Parker Schuh and
                  Kensen Shi and
                  Sasha Tsvyashchenko and
                  Joshua Maynez and
                  Abhishek Rao and
                  Parker Barnes and
                  Yi Tay and
                  Noam Shazeer and
                  Vinodkumar Prabhakaran and
                  Emily Reif and
                  Nan Du and
                  Ben Hutchinson and
                  Reiner Pope and
                  James Bradbury and
                  Jacob Austin and
                  Michael Isard and
                  Guy Gur{-}Ari and
                  Pengcheng Yin and
                  Toju Duke and
                  Anselm Levskaya and
                  Sanjay Ghemawat and
                  Sunipa Dev and
                  Henryk Michalewski and
                  Xavier Garcia and
                  Vedant Misra and
                  Kevin Robinson and
                  Liam Fedus and
                  Denny Zhou and
                  Daphne Ippolito and
                  David Luan and
                  Hyeontaek Lim and
                  Barret Zoph and
                  Alexander Spiridonov and
                  Ryan Sepassi and
                  David Dohan and
                  Shivani Agrawal and
                  Mark Omernick and
                  Andrew M. Dai and
                  Thanumalayan Sankaranarayana Pillai and
                  Marie Pellat and
                  Aitor Lewkowycz and
                  Erica Moreira and
                  Rewon Child and
                  Oleksandr Polozov and
                  Katherine Lee and
                  Zongwei Zhou and
                  Xuezhi Wang and
                  Brennan Saeta and
                  Mark Diaz and
                  Orhan Firat and
                  Michele Catasta and
                  Jason Wei and
                  Kathy Meier{-}Hellstern and
                  Douglas Eck and
                  Jeff Dean and
                  Slav Petrov and
                  Noah Fiedel},
  title        = {PaLM: Scaling Language Modeling with Pathways},
  journal      = {J. Mach. Learn. Res.},
  volume       = {24},
  pages        = {240:1--240:113},
  year         = {2023}
}

@article{DBLP:journals/corr/abs-2303-08774,
  author       = {OpenAI},
  title        = {{GPT-4} Technical Report},
  journal      = {CoRR},
  volume       = {abs/2303.08774},
  year         = {2023}
}

@inproceedings{DBLP:conf/icml/DaoG24,
  author       = {Tri Dao and
                  Albert Gu},
  title        = {Transformers are SSMs: Generalized Models and Efficient Algorithms
                  Through Structured State Space Duality},
  booktitle    = {{ICML}},
  publisher    = {OpenReview.net},
  year         = {2024}
}

@article{zhuang2025vargpt,
  title={VARGPT-v1.1: Improve Visual Autoregressive Large Unified Model via Iterative Instruction Tuning and Reinforcement Learning},
  author={Zhuang, Xianwei and Xie, Yuxin and Deng, Yufan and Yang, Dongchao and Liang, Liming and Ru, Jinghan and Yin, Yuguo and Zou, Yuexian},
  journal={arXiv preprint arXiv:2504.02949},
  year={2025}
}

@article{DBLP:journals/corr/abs-2502-19634,
  author       = {Jiazhen Pan and
                  Che Liu and
                  Junde Wu and
                  Fenglin Liu and
                  Jiayuan Zhu and
                  Hongwei Bran Li and
                  Chen Chen and
                  Cheng Ouyang and
                  Daniel Rueckert},
  title        = {MedVLM-R1: Incentivizing Medical Reasoning Capability of Vision-Language
                  Models (VLMs) via Reinforcement Learning},
  journal      = {CoRR},
  volume       = {abs/2502.19634},
  year         = {2025}
}

@article{DBLP:journals/ijcv/TangJXL25,
  author       = {Lv Tang and
                  Peng{-}Tao Jiang and
                  Haoke Xiao and
                  Bo Li},
  title        = {Towards Training-Free Open-World Segmentation via Image Prompt Foundation
                  Models},
  journal      = {Int. J. Comput. Vis.},
  volume       = {133},
  number       = {1},
  pages        = {1--15},
  year         = {2025}
}

@inproceedings{DBLP:conf/cvpr/LaiTCLY0J24,
  author       = {Xin Lai and
                  Zhuotao Tian and
                  Yukang Chen and
                  Yanwei Li and
                  Yuhui Yuan and
                  Shu Liu and
                  Jiaya Jia},
  title        = {{LISA:} Reasoning Segmentation via Large Language Model},
  booktitle    = {{CVPR}},
  pages        = {9579--9589},
  publisher    = {{IEEE}},
  year         = {2024}
}

@article{DBLP:journals/corr/abs-2304-07193,
  author       = {Maxime Oquab and
                  Timoth{\'{e}}e Darcet and
                  Th{\'{e}}o Moutakanni and
                  Huy Vo and
                  Marc Szafraniec and
                  Vasil Khalidov and
                  Pierre Fernandez and
                  Daniel Haziza and
                  Francisco Massa and
                  Alaaeldin El{-}Nouby and
                  Mahmoud Assran and
                  Nicolas Ballas and
                  Wojciech Galuba and
                  Russell Howes and
                  Po{-}Yao Huang and
                  Shang{-}Wen Li and
                  Ishan Misra and
                  Michael G. Rabbat and
                  Vasu Sharma and
                  Gabriel Synnaeve and
                  Hu Xu and
                  Herv{\'{e}} J{\'{e}}gou and
                  Julien Mairal and
                  Patrick Labatut and
                  Armand Joulin and
                  Piotr Bojanowski},
  title        = {DINOv2: Learning Robust Visual Features without Supervision},
  journal      = {CoRR},
  volume       = {abs/2304.07193},
  year         = {2023}
}

@inproceedings{DBLP:conf/icml/RadfordKHRGASAM21,
  author       = {Alec Radford and
                  Jong Wook Kim and
                  Chris Hallacy and
                  Aditya Ramesh and
                  Gabriel Goh and
                  Sandhini Agarwal and
                  Girish Sastry and
                  Amanda Askell and
                  Pamela Mishkin and
                  Jack Clark and
                  Gretchen Krueger and
                  Ilya Sutskever},
  title        = {Learning Transferable Visual Models From Natural Language Supervision},
  booktitle    = {{ICML}},
  volume       = {139},
  pages        = {8748--8763},
  publisher    = {{PMLR}},
  year         = {2021}
}

@inproceedings{DBLP:conf/icml/0008LSH23,
  author       = {Junnan Li and
                  Dongxu Li and
                  Silvio Savarese and
                  Steven C. H. Hoi},
  title        = {{BLIP-2:} Bootstrapping Language-Image Pre-training with Frozen Image
                  Encoders and Large Language Models},
  booktitle    = {{ICML}},
  volume       = {202},
  pages        = {19730--19742},
  year         = {2023}
}

@inproceedings{DBLP:conf/nips/ShiDX24,
  author       = {Yuheng Shi and
                  Minjing Dong and
                  Chang Xu},
  title        = {Multi-Scale VMamba: Hierarchy in Hierarchy Visual State Space Model},
  booktitle    = {NeurIPS},
  year         = {2024}
}

@article{DBLP:journals/corr/abs-2405-14858,
  author       = {Feng Wang and
                  Jiahao Wang and
                  Sucheng Ren and
                  Guoyizhe Wei and
                  Jieru Mei and
                  Wei Shao and
                  Yuyin Zhou and
                  Alan L. Yuille and
                  Cihang Xie},
  title        = {Mamba-R: Vision Mamba {ALSO} Needs Registers},
  journal      = {CoRR},
  volume       = {abs/2405.14858},
  year         = {2024}
}

@inproceedings{DBLP:conf/aaai/0008ZZDHW25,
  author       = {Han Zhao and
                  Min Zhang and
                  Wei Zhao and
                  Pengxiang Ding and
                  Siteng Huang and
                  Donglin Wang},
  title        = {Cobra: Extending Mamba to Multi-Modal Large Language Model for Efficient
                  Inference},
  booktitle    = {{AAAI}},
  pages        = {10421--10429},
  publisher    = {{AAAI} Press},
  year         = {2025}
}

@article{DBLP:journals/corr/abs-2408-12245,
  author       = {Haopeng Li and
                  Jinyue Yang and
                  Kexin Wang and
                  Xuerui Qiu and
                  Yuhong Chou and
                  Xin Li and
                  Guoqi Li},
  title        = {Scalable Autoregressive Image Generation with Mamba},
  journal      = {CoRR},
  volume       = {abs/2408.12245},
  year         = {2024}
}

@article{DBLP:journals/pami/LiJHWZLMZ25,
  author       = {Yunxin Li and
                  Shenyuan Jiang and
                  Baotian Hu and
                  Longyue Wang and
                  Wanqi Zhong and
                  Wenhan Luo and
                  Lin Ma and
                  Min Zhang},
  title        = {Uni-MoE: Scaling Unified Multimodal LLMs With Mixture of Experts},
  journal      = {{IEEE} Trans. Pattern Anal. Mach. Intell.},
  volume       = {47},
  number       = {5},
  pages        = {3424--3439},
  year         = {2025}
}

@article{liu2025graph,
  title={Graph Foundation Models: Concepts, Opportunities and Challenges},
  author={Liu, Jiawei and Yang, Cheng and Lu, Zhiyuan and Chen, Junze and Li, Yibo and Zhang, Mengmei and Bai, Ting and Fang, Yuan and Sun, Lichao and Yu, Philip S and others},
  journal={IEEE Transactions on Pattern Analysis and Machine Intelligence},
  year={2025},
  publisher={IEEE}
}

@article{awais2025foundation,
  title={Foundation Models Defining a New Era in Vision: a Survey and Outlook},
  author={Awais, Muhammad and Naseer, Muzammal and Khan, Salman and Anwer, Rao Muhammad and Cholakkal, Hisham and Shah, Mubarak and Yang, Ming-Hsuan and Khan, Fahad Shahbaz},
  journal={IEEE Transactions on Pattern Analysis and Machine Intelligence},
  year={2025},
  publisher={IEEE}
}

@article{zhou2019semantic,
  title={Semantic understanding of scenes through the ade20k dataset},
  author={Zhou, Bolei and Zhao, Hang and Puig, Xavier and Xiao, Tete and Fidler, Sanja and Barriuso, Adela and Torralba, Antonio},
  journal={International Journal of Computer Vision},
  volume={127},
  number={3},
  pages={302--321},
  year={2019},
  publisher={Springer}
}

@article{DBLP:journals/ijon/SuALPBL24,
  author       = {Jianlin Su and
                  Murtadha H. M. Ahmed and
                  Yu Lu and
                  Shengfeng Pan and
                  Wen Bo and
                  Yunfeng Liu},
  title        = {RoFormer: Enhanced transformer with Rotary Position Embedding},
  journal      = {Neurocomputing},
  volume       = {568},
  pages        = {127063},
  year         = {2024}
}

@article{DBLP:journals/corr/abs-2407-08083,
  author       = {Ali Hatamizadeh and
                  Jan Kautz},
  title        = {MambaVision: {A} Hybrid Mamba-Transformer Vision Backbone},
  journal      = {CoRR},
  volume       = {abs/2407.08083},
  year         = {2024}
}

@article{DBLP:journals/corr/abs-2410-15091,
  author       = {Chaodong Xiao and
                  Minghan Li and
                  Zhengqiang Zhang and
                  Deyu Meng and
                  Lei Zhang},
  title        = {Spatial-Mamba: Effective Visual State Space Models via Structure-Aware State Fusion},
  journal      = {CoRR},
  volume       = {abs/2410.15091},
  year         = {2024}
}

@inproceedings{DBLP:conf/nips/HanWXHPGSSZ024,
  author       = {Dongchen Han and
                  Ziyi Wang and
                  Zhuofan Xia and
                  Yizeng Han and
                  Yifan Pu and
                  Chunjiang Ge and
                  Jun Song and
                  Shiji Song and
                  Bo Zheng and
                  Gao Huang},
  title        = {Demystify Mamba in Vision: {A} Linear Attention Perspective},
  booktitle    = {NeurIPS},
  year         = {2024}
}

@article{DBLP:journals/corr/abs-2406-02395,
  author       = {Yicheng Xiao and
                  Lin Song and
                  Shaoli Huang and
                  Jiangshan Wang and
                  Siyu Song and
                  Yixiao Ge and
                  Xiu Li and
                  Ying Shan},
  title        = {GrootVL: Tree Topology is All You Need in State Space Model},
  journal      = {CoRR},
  volume       = {abs/2406.02395},
  year         = {2024}
}

@inproceedings{DBLP:conf/nips/LiuTZYX0YJ024,
  author       = {Yue Liu and
                  Yunjie Tian and
                  Yuzhong Zhao and
                  Hongtian Yu and
                  Lingxi Xie and
                  Yaowei Wang and
                  Qixiang Ye and
                  Jianbin Jiao and
                  Yunfan Liu},
  title        = {VMamba: Visual State Space Model},
  booktitle    = {NeurIPS},
  year         = {2024}
}

@inproceedings{DBLP:conf/icml/ZhuL0W0W24,
  author       = {Lianghui Zhu and
                  Bencheng Liao and
                  Qian Zhang and
                  Xinlong Wang and
                  Wenyu Liu and
                  Xinggang Wang},
  title        = {Vision Mamba: Efficient Visual Representation Learning with Bidirectional
                  State Space Model},
  booktitle    = {{ICML}},
  publisher    = {OpenReview.net},
  year         = {2024}
}

@article{liu2025defmamba,
  title={DefMamba: Deformable Visual State Space Model},
  author={Liu, Leiye and Zhang, Miao and Yin, Jihao and Liu, Tingwei and Ji, Wei and Piao, Yongri and Lu, Huchuan},
  journal={arXiv preprint arXiv:2504.05794},
  year={2025}
}

@inproceedings{DBLP:conf/aaai/Pei0X25,
  author       = {Xiaohuan Pei and
                  Tao Huang and
                  Chang Xu},
  title        = {EfficientVMamba: Atrous Selective Scan for Light Weight Visual Mamba},
  booktitle    = {{AAAI}},
  pages        = {6443--6451},
  publisher    = {{AAAI} Press},
  year         = {2025}
}

@inproceedings{xiao2018unified,
  title={Unified perceptual parsing for scene understanding},
  author={Xiao, Tete and Liu, Yingcheng and Zhou, Bolei and Jiang, Yuning and Sun, Jian},
  booktitle={{ECCV}},
  pages={418--434},
  year={2018}
}

@article{chen2019mmdetection,
  title={MMDetection: Open mmlab detection toolbox and benchmark},
  author={Chen, Kai and Wang, Jiaqi and Pang, Jiangmiao and Cao, Yuhang and Xiong, Yu and Li, Xiaoxiao and Sun, Shuyang and Feng, Wansen and Liu, Ziwei and Xu, Jiarui and others},
  journal={arXiv preprint arXiv:1906.07155},
  year={2019}
}

@inproceedings{lin2014microsoft,
  title={Microsoft coco: Common objects in context},
  author={Lin, Tsung-Yi and Maire, Michael and Belongie, Serge and Hays, James and Perona, Pietro and Ramanan, Deva and Doll{\'a}r, Piotr and Zitnick, C Lawrence},
  booktitle={{ECCV}},
  pages={740--755},
  year={2014},
  organization={Springer}
}

@article{loshchilov2017decoupled,
  title={Decoupled weight decay regularization},
  author={Loshchilov, Ilya and Hutter, Frank},
  journal={arXiv preprint arXiv:1711.05101},
  year={2017}
}

@inproceedings{deng2009imagenet,
  title={Imagenet: A large-scale hierarchical image database},
  author={Deng, Jia and Dong, Wei and Socher, Richard and Li, Li-Jia and Li, Kai and Fei-Fei, Li},
  booktitle={{CVPR}},
  pages={248--255},
  year={2009},
  organization={IEEE}
}

@inproceedings{DBLP:conf/iclr/GuGR22,
  author       = {Albert Gu and
                  Karan Goel and
                  Christopher R{\'{e}}},
  title        = {Efficiently Modeling Long Sequences with Structured State Spaces},
  booktitle    = {{ICLR}},
  publisher    = {OpenReview.net},
  year         = {2022}
}

@inproceedings{DBLP:conf/iccv/LiuL00W0LG21,
  author       = {Ze Liu and
                  Yutong Lin and
                  Yue Cao and
                  Han Hu and
                  Yixuan Wei and
                  Zheng Zhang and
                  Stephen Lin and
                  Baining Guo},
  title        = {Swin Transformer: Hierarchical Vision Transformer using Shifted Windows},
  booktitle    = {{ICCV}},
  pages        = {9992--10002},
  publisher    = {{IEEE}},
  year         = {2021}
}

@inproceedings{DBLP:conf/eccv/TouvronCJ22,
  author       = {Hugo Touvron and
                  Matthieu Cord and
                  Herv{\'{e}} J{\'{e}}gou},
  title        = {DeiT {III:} Revenge of the ViT},
  booktitle    = {{ECCV}},
  volume       = {13684},
  pages        = {516--533},
  publisher    = {Springer},
  year         = {2022}
}

@inproceedings{DBLP:conf/iclr/DosovitskiyB0WZ21,
  author       = {Alexey Dosovitskiy and
                  Lucas Beyer and
                  Alexander Kolesnikov and
                  Dirk Weissenborn and
                  Xiaohua Zhai and
                  Thomas Unterthiner and
                  Mostafa Dehghani and
                  Matthias Minderer and
                  Georg Heigold and
                  Sylvain Gelly and
                  Jakob Uszkoreit and
                  Neil Houlsby},
  title        = {An Image is Worth 16x16 Words: Transformers for Image Recognition
                  at Scale},
  booktitle    = {{ICLR}},
  publisher    = {OpenReview.net},
  year         = {2021}
}

@article{DBLP:journals/corr/HowardZCKWWAA17,
  author       = {Andrew G. Howard and
                  Menglong Zhu and
                  Bo Chen and
                  Dmitry Kalenichenko and
                  Weijun Wang and
                  Tobias Weyand and
                  Marco Andreetto and
                  Hartwig Adam},
  title        = {MobileNets: Efficient Convolutional Neural Networks for Mobile Vision
                  Applications},
  journal      = {CoRR},
  volume       = {abs/1704.04861},
  year         = {2017}
}

@inproceedings{DBLP:journals/corr/SimonyanZ14a,
  author       = {Karen Simonyan and
                  Andrew Zisserman},
  title        = {Very Deep Convolutional Networks for Large-Scale Image Recognition},
  booktitle    = {{ICLR}},
  year         = {2015}
}

@inproceedings{DBLP:conf/nips/KrizhevskySH12,
  author       = {Alex Krizhevsky and
                  Ilya Sutskever and
                  Geoffrey E. Hinton},
  title        = {ImageNet Classification with Deep Convolutional Neural Networks},
  booktitle    = {{NIPS}},
  pages        = {1106--1114},
  year         = {2012}
}

@inproceedings{DBLP:conf/cvpr/HeZRS16,
  author       = {Kaiming He and
                  Xiangyu Zhang and
                  Shaoqing Ren and
                  Jian Sun},
  title        = {Deep Residual Learning for Image Recognition},
  booktitle    = {{CVPR}},
  pages        = {770--778},
  publisher    = {{IEEE}},
  year         = {2016}
}

@article{DBLP:journals/corr/abs-2403-17695,
  author       = {Chenhongyi Yang and
                  Zehui Chen and
                  Miguel Espinosa and
                  Linus Ericsson and
                  Zhenyu Wang and
                  Jiaming Liu and
                  Elliot J. Crowley},
  title        = {PlainMamba: Improving Non-Hierarchical Mamba in Visual Recognition},
  journal      = {CoRR},
  volume       = {abs/2403.17695},
  year         = {2024}
}

@article{DBLP:journals/corr/abs-2403-09338,
  author       = {Tao Huang and
                  Xiaohuan Pei and
                  Shan You and
                  Fei Wang and
                  Chen Qian and
                  Chang Xu},
  title        = {LocalMamba: Visual State Space Model with Windowed Selective Scan},
  journal      = {CoRR},
  volume       = {abs/2403.09338},
  year         = {2024}
}

@article{DBLP:journals/corr/abs-2312-00752,
  author       = {Albert Gu and
                  Tri Dao},
  title        = {Mamba: Linear-Time Sequence Modeling with Selective State Spaces},
  journal      = {CoRR},
  volume       = {abs/2312.00752},
  year         = {2023}
}

@InProceedings{Kirillov_2023_ICCV,
    author    = {Kirillov, Alexander and Mintun, Eric and Ravi, Nikhila and Mao, Hanzi and Rolland, Chloe and Gustafson, Laura and Xiao, Tete and Whitehead, Spencer and Berg, Alexander C. and Lo, Wan-Yen and Dollar, Piotr and Girshick, Ross},
    title     = {Segment Anything},
    booktitle = {{ICCV}},
    month     = {October},
    year      = {2023},
    pages     = {4015-4026}
}

@inproceedings{DBLP:conf/nips/VaswaniSPUJGKP17,
  author       = {Ashish Vaswani and
                  Noam Shazeer and
                  Niki Parmar and
                  Jakob Uszkoreit and
                  Llion Jones and
                  Aidan N. Gomez and
                  Lukasz Kaiser and
                  Illia Polosukhin},
  title        = {Attention is All you Need},
  booktitle    = {{NeurIPS}},
  pages        = {5998--6008},
  year         = {2017}
}

@inproceedings{DBLP:conf/icml/0001LXH22,
  author       = {Junnan Li and
                  Dongxu Li and
                  Caiming Xiong and
                  Steven C. H. Hoi},
  title        = {{BLIP:} Bootstrapping Language-Image Pre-training for Unified Vision-Language
                  Understanding and Generation},
  booktitle    = {{ICML}},
  series       = {Proceedings of Machine Learning Research},
  volume       = {162},
  pages        = {12888--12900},
  publisher    = {{PMLR}},
  year         = {2022}
}

@article{DBLP:journals/corr/abs-2302-13971,
  author       = {Hugo Touvron and
                  Thibaut Lavril and
                  Gautier Izacard and
                  Xavier Martinet and
                  Marie{-}Anne Lachaux and
                  Timoth{\'{e}}e Lacroix and
                  Baptiste Rozi{\`{e}}re and
                  Naman Goyal and
                  Eric Hambro and
                  Faisal Azhar and
                  Aur{\'{e}}lien Rodriguez and
                  Armand Joulin and
                  Edouard Grave and
                  Guillaume Lample},
  title        = {LLaMA: Open and Efficient Foundation Language Models},
  journal      = {CoRR},
  volume       = {abs/2302.13971},
  year         = {2023}
}

@article{chen2404changemamba,
  title={ChangeMamba: Remote Sensing Change Detection with Spatio-Temporal State Space Model. arXiv 2024},
  author={Chen, H and Song, J and Han, C and Xia, J and Yokoya, N},
  journal={arXiv preprint arXiv:2404.03425},
  year={2024}
}

@inproceedings{DBLP:conf/eccv/ZhengJWZCWL24,
  author       = {Tianyi Zheng and
                  Peng{-}Tao Jiang and
                  Ben Wan and
                  Hao Zhang and
                  Jinwei Chen and
                  Jia Wang and
                  Bo Li},
  title        = {Beta-Tuned Timestep Diffusion Model},
  booktitle    = {{ECCV} {(3)}},
  series       = {Lecture Notes in Computer Science},
  volume       = {15061},
  pages        = {114--130},
  publisher    = {Springer},
  year         = {2024}
}
}

\end{document}